# Two Steps Feature Selection and Neural Network Classification for the TREC-8 Routing

Mathieu Stricker [*,**]      Frantz Vichot [*]      Gérard Dreyfus [**]      Francis Wolinski [*]

[*] Informatique-CDC – Groupe Caisse des Dépôts
Direction des Techniques Avancées
4 rue Berthollet
94114 Arcueil cedex - France
{forename.surname}@icdc.caissedesdepots.fr

[**] ESPCI
Laboratoire d'Electronique
10 rue Vauquelin
75005 Paris, France
{forename.surname}@espci.fr

## 1   Introduction

At the Caisse des Dépôts et Consignations (CDC), the Agence France-Presse (AFP) news releases are filtered continuously according to the users' interests. Once a user has specified a topic of interest, a filter is customized to fit this user's profile. Until now, these filters would rely on rule-based methods, whose efficiency is proven [Vichot *et al.*, 1999], but which require a large amount of work for each specific filter. This drawback can be avoided by using statistical methods which have the ability to learn from examples of relevant documents. Recently, we have developed a methodology for the AFP corpus. This paper presents its application to the TREC-8 corpus.

For the TREC-8 routing, one specific filter is built for each topic. Each filter is a classifier trained to recognize the documents that are relevant to the topic. When presented with a document, each classifier estimates the probability for the document to be relevant to the topic for which it has been trained. Since the procedure for building a filter is topic-independent, the system is fully automatic. Therefore, we describe it for one topic; the procedure is repeated 50 times.

By making use of a sample of documents that have previously been evaluated as relevant or not relevant to a particular topic, a term selection is performed, and a neural network is trained. Each document is represented by a vector of frequencies of a list of selected terms. This list depends on the topic to be filtered; it is constructed in two steps. The first step defines the characteristic words used in the relevant documents of the corpus; the second one chooses, among the previous list, the most discriminant ones. The length of the vector is optimized automatically for each topic. At the end of the term selection, a vector of typically 25 words is defined for the topic, so that each document which has to be processed is represented by a vector of term frequencies.

This vector is subsequently input to a classifier that is trained from the same sample. After training, the classifier estimates for each document of a test set its probability of being relevant; for submission to TREC, the top 1000 documents are ranked in order of decreasing relevance.

## 2   Problem and data description

In order to build the users' profile, routing participants were allowed to use the relevance judgments from the 1992 Financial Times (FT92) collection and any other relevance judgments from any other parts of the TREC collection except the 1993-94 Financial Times collection (FT93-94).

The FT93-94 collection is the test set; it contains 140,650 documents. Participants are asked to return a ranked list of the top 1000 retrieved documents from this part of the collection for each topic.

The routing profiles for TREC-8 are to be built for topics 351 to 400. Relevance judgments for these topics are provided on the FT92 collection. Since the test set is part of the FT collection, it would have been desirable to use only the FT92 collection for training. However, there are, on average, only 11 relevant documents per topic on the FT92 collection. Since this number is too low for estimating safely the parameters of a statistical model, the training set, in our experiments, includes all the relevant documents available for these topics on the whole TREC corpus (except FT93-94): the Federal Register 1994 collection (FR94), the Foreign Broadcast Information Service (FBIS) collection and the LA Times collection. With these additional collections, the average number of relevant documents per topic is increased to 71.6 (median 55.5).

## 3   Building of the training set

In order to build an efficient filter, the training set must be large and representative enough of the classes to be learned. A subset of relevant examples and a subset of irrelevant ones compose the training set. Since the number of relevant documents for each topic is generally low, all the available relevant documents from FT92, FBIS, LA and FR are selected.

For the subset of irrelevant documents, the number of candidates is huge. However, since the subset of relevant documents typically includes less than one hundred examples, the subset of irrelevant documents is limited to a few thousands. For each topic, two categories of irrelevant documents are available:
1. Those which have been judged irrelevant by a relevance assessor of TREC.
2. Those which have never been looked at; they are assumed to be irrelevant.

The documents from the first category have been checked because they were suspected to be relevant by some previous system (see TREC overview papers [Voorhees and Harman, 1999] for more details on the pooling technique used in TREC). Therefore, these documents can be said to be 'close' to the relevant documents: they are not representatives of the class of the irrelevant documents.

Some first experiments have shown that the results are better if we consider only irrelevant documents from the second category. Consequently, we sampled randomly 3,000 documents from this category only.

# 4  Term Selection

Each document of the FT92 collection is first tokenized into single words, case being ignored. In the following, each word is considered as a single unit called term. No stemming is performed.

We use no controlled vocabulary fields from the FT collection for building the filters.

The goal of the term selection is to define, for each topic, a vector of terms that will represent each document. The choice of these terms must be done very carefully since the quality of the filter relies heavily on this choice, whatever the model is. These terms must be chosen to allow a classifier to discriminate between relevant and irrelevant documents. The number of these terms is a tradeoff between two requirements: the larger the number of terms, the larger the number of examples required to have a good estimate of the classifier parameters; however, discarding terms leads to information loss.

Consequently a term selection must choose an appropriate number of appropriate terms for each topic.

## 4.1  Topic frequency analysis

The total frequency of each term of the FT92 collection is computed. This value is called corpus term frequency.

For each term of a relevant document, we compute the document term frequency divided by the corpus term frequency. The terms of the document are then sorted by decreasing order according to this ratio. Therefore, very common words (stop words), which have a very high frequency in the whole corpus, are at the bottom of the list. The most specific words are at the top of the list, and the very rare words are ranked first.

The first half of the list is saved and the second half is discarded for each document. All the lists of all the relevant documents are merged and the frequencies of each single term in this list are computed.

A final sorting of the terms is performed, in order of decreasing frequencies, so that the rare words are at the bottom of the list.

At the end of this process we have a list of terms from which very common terms and very rare terms have been discarded; the remaining terms are representative of the specific vocabulary of the topic.

### Heterogeneity of the training set

The subset of relevant documents in the training set arises from several collections, but the corpus term frequencies are computed on the FT92 collection. Consequently, some words are much less frequent on the FT92 collection than on other corpuses. For example, the word "california" tends to occur much more frequently on the LA Times collection than on the FT92 collection.

Consequently a short list of stop words was defined to take into account this heterogeneity. This list includes words like "california", "los", "angeles", … . A better solution might have been to compute the frequencies over the whole collection, but since the test set was only part of FT collection, the solution with the stop words list was adopted.

This first step defines the vocabulary specific to the topic, but the remaining terms are not necessarily the most discriminant ones and some of them may be highly correlated like "buenos" and "aires" for topic 351.

The goal of the second part of the term selection is to choose, amongst the previous words, the most discriminant ones in order to achieve a good classification task.

## 4.2  Gram-Schmidt

The Gram-Schmidt orthogonalization technique [Chen *et al.*, 1989] is used to rank the remaining terms in order of decreasing relevance to the output. The method can be described as follows.

We consider a model with $Q$ candidate terms, and a training set containing $N$ examples of documents whose relevance is known. The relevance of each document is considered as the desired output of the model: +1 for a relevant document and –1 for an irrelevant document. We denote by $x^i = {}^T[tf_1^i, tf_2^i, ..., tf_N^i]$ the vector of term frequencies of term $i$ in the different examples. We denote by $y_p$ the $N$-vector of the outputs to be modeled. We consider the $(N, Q)$ matrix $X = [\mathbf{tf}^1, \mathbf{tf}^2, ..., \mathbf{tf}^Q]$. The model can be written as $y = X\vartheta$, where $\vartheta$ is the vector of the parameters of the model.

The first iteration of the procedure consists in finding the vector of terms which best explains the output, i.e. which has the smallest angle with the output vector in the *N*-dimensional space of observations. To this end, the following quantities are computed:

$$\cos^2\{x^k, y_p\} = \frac{\{^Tx^k y_p\}^2}{\{^Tx^k x^k\}\{^Ty_p y_p\}} \;,\; k = 1 \text{ to } Q$$

and the vector for which this quantity is largest is selected. In order to eliminate the part of the output which is explained by the first selected vector, all remaining candidate inputs, and the output vector, are projected onto the null subspace (of dimension *N*-1) of the selected term. In this subspace, the projected input vector that best explains the projected output is selected, and the *Q*-2 remaining vectors of terms are projected onto the null subspace of the first two ranked vectors. The procedure terminates when all *Q* input vectors are ranked. At the end of this procedure, a list of terms, ranked in order of decreasing relevance, is available.

This method applies only to models that are linear with respect to their parameters. This drawback can be circumvented by noting that an input, that is irrelevant for a model linear with respect to its parameters, is very likely to be irrelevant, irrespective of the model. Therefore, term selection is performed with Gram-Schmidt orthogonalization, and the selected terms are used as inputs to a neural network.

Once the terms are ranked, the pending question is that of deciding to what depth, within the list, the terms should be selected. This is achieved by introducing a "probe" term, which is a random variable, and by ranking this term, just as all other candidate terms, with the procedure described in the above section. The candidate terms that are ranked below the probe should be discarded. Actually, the rank of the probe term is a random variable; therefore, one has to compute the cumulative distribution function of this variable, and, in the spirit of hypothesis tests, one must choose a risk of selecting a term although it is less relevant than a random input (typically 1% or 5%). The computation of the probability for a probe to be more significant than one of the *n* terms selected after iteration *n* can be found in [Stoppiglia, 1997].

### 4.3 TREC-8 Result for term selection

The term selection method described in the previous section generates a vector of discriminant terms automatically for each topic. The dimension of this vector is determined by the probe defined above and thus its length is customized for each topic. The average length of the vector over all the topics is 25 terms, the maximum length is 40, and the minimum length is 10.

Figure 2 shows the final 10 top terms of topics 351, 352 and 375:

| Topic 351    | Topic 352    | Topic 375   |
|--------------|--------------|-------------|
| islands      | channel      | fusion      |
| aires        | rail         | fuel        |
| argentine    | terminal     | energy      |
| carlos       | link         | science     |
| argentina    | kent         | hydrogen    |
| exploration  | eurotunnel   | produced    |
| falkland     | developments | reaction    |
| falklands    | tunnel       | cold        |
| menem        | jobs         | electric    |
| exploitation | railway      | laboratories |

Figure 1: Final 10 top words of the term selection

## 5 Neural Networks

For each topic, the term frequencies of the vectors defined above are used as inputs of a neural network. Since the number of relevant examples per topic is low, the simplest architecture is chosen for the neural network i.e. a simple unit with a hyperbolic tangent function. Therefore, our classifier actually performs a linear separation.

Each filter is the function: $\tanh\left(\sum_{i=0}^{N} w_i x_i\right)$

Where N is the size of the input vector for the topic considered, $w_i$ are the parameters to be estimated called weights and $x_i$ is defined for i>0 by:

$$x_i = \begin{cases} -1 & tf_i = 0 \\ \dfrac{tf_i}{Log(L)} & tf_i > 0 \end{cases}$$

where $tf_i$ are the term frequencies of the terms selected by the term selection procedure. For each classifier $x_0 = 1$. L is the length of the text. By dividing each term frequency by the logarithm of the length, we take into account the fact that for longer text, the terms tends to appear more often.

The classifier is trained by minimizing the mean square difference between the desired output and the actual value of the classifier on the training set. The desired value is +1 if the document is relevant and –1 if the document is not relevant.

After training, each document of the test set (FT93-94 collection) is processed through the network and the output of the network is a number between [-1;+1]. The document are then sorted by decreasing order of output, and the top 1,000 documents are submitted for evaluation.

There are several techniques for minimizing the mean square error; one of the most efficient algorithm is the BFGS Quasi-Newton algorithm [Bishop, 1995]. However, since the number of relevant examples is small, the minimum of the cost function corresponds to large weights, so that the networks produce essentially a binary output: +1 or –1. In this case, the network has been overtrained, and is generalization ability is low. Consequently, in order to avoid the saturation of the tanh function, an early stopping procedure is used: the cost is minimized with a gradient descent procedure, and training is stopped after a few epochs.

In other experiments we have tried to use hidden neurons in the network architecture, but it leads to no improvement in the results.

## 6   Result on the FT93-94 collection

Our methodology has been applied to the topics 351 to 400 for the routing subtasks. The performances are measured by uninterpolated average precision.

The results produced by the trec_aval[1] package are listed below. The topics 359 and 369, which have no relevant document in FT93-94, are not taken into account.

```
Retrieved:                                  48000
Relevant:                                    1276
Relevant Retrieved:                          1129

Non interpolated average precision :        31,99
R-precision                                 31,59
```

| Interpolated Recall | Precision Averages |
|---|---|
| 0 | 67,69 |
| 10 | 56,1 |
| 20 | 48,98 |
| 30 | 41,84 |
| 40 | 37,39 |
| 50 | 32,5 |
| 60 | 25,61 |
| 70 | 21,54 |
| 80 | 18,73 |
| 90 | 13,51 |
| 100 | 11,36 |

| Precision at | | |
|---|---|---|
| 5 | docs | 40,00 |
| 10 | docs | 31,67 |
| 15 | docs | 27,50 |
| 20 | docs | 24,90 |
| 30 | docs | 21,18 |
| 100 | docs | 12,27 |
| 200 | docs | 8,10 |
| 500 | docs | 4,19 |
| 1000 | docs | 2,35 |

---

[1] Available at ftp://ftp.cs.cornell.edu/pub/smart/

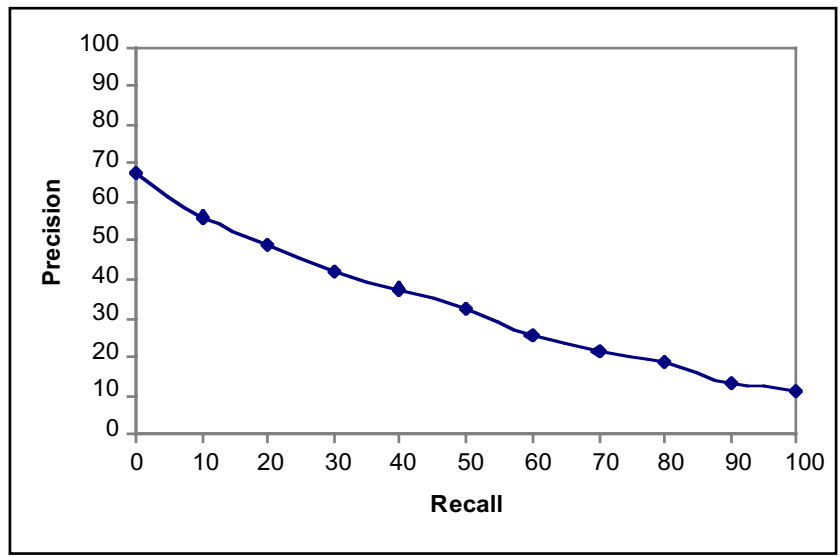

## 7 Future Work

In this work, the terms used as inputs of the classifiers were actually single words. In the future, we plan to improve the text representation by using N-grams. These new terms can be handle in our method exactly like the single words. We expect the use of these N-grams to improve slightly the overall performance.

Our classifier was just a logistic regression due to the lack of training data available and the risk of overfitting. An early stopping procedure was used during the training process, but the number of epochs was not optimized for each topic. In some future experiments, we plan to use a weight decay regularizers instead of the early stopping procedure. The weight decay consists of the sum of squares of the adaptive parameters in the network: a new cost function $E'$ has to be to minimized instead of $E$ the mean square error between the desired output and the actual value of the network.

$$E' = E + \frac{\lambda}{2} \sum_{i=1}^{n} w_i^2$$

Where $w_i$ are the adaptive weights of the networks. $\lambda$ is a new parameter which value has to be determined. This approach has been studied intensively in [MacKay, 1992] and seems to be promising.